\documentclass{article}

\PassOptionsToPackage{numbers, compress}{natbib}
\usepackage[final]{neurips_2022}

\usepackage[utf8]{inputenc} 
\usepackage[T1]{fontenc}    
\usepackage{hyperref}       
\usepackage{url}            
\usepackage{booktabs}       
\usepackage{amsfonts}       
\usepackage{nicefrac}       
\usepackage{microtype}      
\usepackage{xcolor}         

\usepackage{amsmath}
\usepackage{subfig}
\usepackage{tikz}
\usetikzlibrary{shapes.geometric, arrows}

\newcommand{\vz}{\mathbf{z}}
\newcommand{\vs}{\mathbf{s}}
\newcommand{\va}{\mathbf{a}}
\newcommand{\vm}{\mathbf{m}}
\newcommand{\vx}{\mathbf{x}}
\newcommand{\mZ}{\mathbf{Z}}
\newcommand{\mA}{\mathbf{A}}
\newcommand{\mC}{\mathbf{C}}

\tikzstyle{text} = [font=\large, text centered]

\tikzstyle{cls} = [rectangle, rounded corners, minimum width=0.4cm, minimum height=0.4cm,text centered, draw=black]

\tikzstyle{patch} = [rectangle, rounded corners, minimum width=0.4cm, minimum height=0.4cm,text centered, draw=black, fill=cyan!30]

\tikzstyle{masked_patch} = [rectangle, rounded corners, minimum width=0.5cm, minimum height=0.5cm,text centered, draw=black, fill=gray!30]

\tikzstyle{dot_product} = [circle, rounded corners, minimum width=0.2cm, minimum height=0.2cm,text centered, draw=black, scale=0.6]

\tikzstyle{function} = [rectangle,rounded corners,text centered, draw=black,rotate=90,minimum width=3cm, minimum height=1cm,fill=gray!30]

\tikzstyle{decoder} = [rectangle, rounded corners, minimum width=1cm, minimum height=2cm,text centered, draw=black, fill=gray!30]

\title{Learning Explicit Object-Centric Representations\\
with Vision Transformers}

\author{
  Oscar Vikström\\
  Aalto University\\
  \texttt{oscar.vikstrom@aalto.fi} \\
  \And
  Alexander Ilin \\
  Aalto University\\
  \texttt{alexander.ilin@aalto.fi} \\
}

\begin{document}

\maketitle

\begin{abstract}
  With the recent successful adaptation of transformers to the vision domain, particularly when trained in a self-supervised fashion, it has been shown that vision transformers can learn impressive object-reasoning-like behaviour and features expressive for the task of object segmentation in images. In this paper, we build on the self-supervision task of masked autoencoding and explore its effectiveness for explicitly learning object-centric representations with transformers. To this end, we design an object-centric autoencoder using transformers only and train it end-to-end to reconstruct full images from unmasked patches. We show that the model efficiently learns to decompose simple scenes as measured by segmentation metrics on several multi-object benchmarks.
\end{abstract}

\section{Introduction}

We consider the problem of self-supervised learning of explicit object-centric representations using the Vision Transformer (ViT) \cite{vit} architecture. In particular, we focus on the task of learning to segment and represent multi-object scenes without using manually annotated labels. We hypothesize that the object-reasoning-like behaviour demonstrated to appear in the self-supervision task of masked image modelling \cite{mae, simmim} can be utilized also for explicit object-centric representation learning. To this end, we investigate using masked autoencoding as the training task for object-centric learning. 

More specifically, we adapt the Masked Autoencoder (MAE) \cite{mae} design and modify it to for explicit object-centric representation learning and segmentation. The architecture is purely based on transformers and we extract object representations in specially reserved positions. Unlike in \cite{mae}, the model is trained to reconstruct the full image as a mixture of individual component reconstructions. Since this task requires both reducing the image to a set of representations, and object reasoning due to partial observation, the model is encouraged to encode meaningful entities that can be reasoned about, like objects or background, with its available representations. The model is trained end-to-end with an initially high masking ratio which is linearly decreased during the training.

We compare our approach using segmentation metrics to state-of-the-art 
methods on common synthetic multi-object datasets and also on the more difficult ClevrTex dataset \cite{clevrtex}. We show that the model trained with masking  efficiently learns object-centric representations that lead to segmentation performance on par with previous methods on the simple multi-object datasets, and while unstable, it shows some promise also on the more difficult ClevrTex data.

The main contributions of the paper are:
the introduction of masked autoencoding for object-centric learning; a method for object representation extraction using multiple class tokens from a transformer encoder; an efficient and fully transformer based object-centric autoencoder.

\section{Related Work}

\textbf{Self-supervised Learning} \ While supervised training of ViTs typically requires pre-training on large datasets, self-supervised training  has allowed for effective pre-training of large models using less data \cite{mocov3, dino, beit, mae, simmim, data2vec}. The self-supervised training of ViTs can lead to interesting object-centric results. As demonstrated in \cite{dino}, the features learned in self-supervised training of the ViT are more clearly indicative of the semantic segmentation in images compared to the ones learned with supervised training. Another example is the object-reasoning-like results seen in the predictions of masked image-modelling frameworks \cite{mae, simmim}. In this work, we capitalize in particular on the masked autoencoding of \cite{mae} and investigate using it for object-centric learning.

\textbf{Object-centric Learning} \ In recent years, there have been a number of methods suggested for unsupervised object-centric learning from images \cite{monet, iodine, slot_attention, genesis, genesisv2, emorl, spair, space, gnm, mn, dti, slate, comet, ast}. These models have been shown to successfully decompose scenes of objects and backgrounds into meaningful object-centric representations, as shown by segmentation metrics, property prediction tasks or capability for compositional generation. However, object-centric learning from complex and varying natural images still remains a great challenge. In fact, as the authors of \cite{clevrtex} demonstrate, many object-centric models struggle when applied to more difficult scenes of textured components. 

\section{Method}
\label{sec:method}

\subsection{Architecture}

\begin{figure}[t]
  \centering
\begin{tikzpicture}[scale=0.65, every node/.style={scale=0.65}]

\node (cls_0) at (1,9) [cls, fill=red!30] {};
\node (cls_1) at (1,8.5) [cls, fill=blue!30] {};
\node (cls_2) at (1,8) [cls, fill=green!30] {};
\node (cls_3) at (1,7.5) [cls, fill=yellow!30] {};

\node (unmasked) [below of=cls_3, anchor=north, node distance=8mm] {\includegraphics[height=110pt]{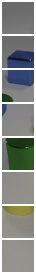}};

\node (input) [left of=unmasked, anchor=east, node distance=10mm] {\includegraphics[width=60pt]{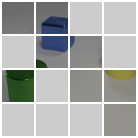}};

\node (encoder) at (2.5,6) [rectangle, rounded corners, minimum width=1cm, minimum width=6.5cm, minimum height=1cm,text centered, draw=black, fill=gray!30,rotate=90] {ENCODER};

\node (cls_0_enc) at (4,9) [cls, fill=red!30] {};
\node (cls_1_enc) at (4,8.5) [cls, fill=blue!30] {};
\node (cls_2_enc) at (4,8) [cls, fill=green!30] {};
\node (cls_3_enc) at (4,7.5) [cls, fill=yellow!30] {};

\node (patch_0_enc) at (4,6.5) [patch] {};
\node (patch_1_enc) at (4,6) [patch] {};
\node (patch_2_enc) at (4,5.5) [patch] {};
\node (patch_3_enc) at (4,5) [patch] {};
\node (patch_4_enc) at (4,4.5) [patch] {};
\node (patch_5_enc) at (4,4) [patch] {};
\node (patch_6_enc) at (4,3.5) [patch] {};
\node (patch_7_enc) at (4,3) [patch] {};

\node (extraction) at (5.5,4.75) [rectangle, rounded corners,text centered, align=center, draw=black, minimum width=4cm,, minimum height=1cm, fill=gray!30, rotate=90] {Object function};

\node (s0) at (7,5.5) [cls, fill=red!30] {};
\node (s1) at (7,5) [cls, fill=blue!30] {};
\node (s2) at (7,4.5) [cls, fill=green!30] {};
\node (s3) at (7,4) [cls, fill=yellow!30] {};

\node (stext) [below of=s3, node distance=5mm] {$\vs$};
\node (atext) [below of=stext] {$\mathbf{a}$};

\node (forming) at (8.5,4.75) [rectangle, rounded corners,text centered, align=center, draw=black, minimum width=4cm,, minimum height=1cm, fill=gray!30, rotate=90] {Broadcasting};

\foreach \Z/\C in {-1.5/yellow,-1/green,-0.5/blue, 0/red}
{
\node (patch_0_mask) at (10,6.5,\Z) [cls, fill=\C!30] {};
\node (patch_1_mask) at (10,6,\Z) [cls, fill=\C!30] {};
\node (patch_2_mask) at (10,5.5,\Z) [cls, fill=gray!30] {};
\node (dots) at (10,4.9,\Z) [] {$\vdots$};
\node (patch_5_mask) at (10,4,\Z) [cls, fill=gray!30] {};
\node (patch_6_mask) at (10,3.5,\Z) [cls, fill=gray!30] {};
\node (patch_7_mask) at (10,3,\Z) [cls, fill=\C!30] {};
}

\foreach \Z in {-1.5,-1,-0.5, 0}
{
\node (decoder) at (12,4.75, \Z) [decoder,rectangle,rounded corners,text centered, draw=black, fill=gray!30,rotate=90,minimum width=4cm,minimum height=1cm] {DECODER};
}

\foreach \Z/\C in {-1.5/yellow,-1/green,-0.5/blue, 0/red}
{
\node (patch_0_mask) at (14,6.5,\Z) [cls, fill=\C!30] {};
\node (patch_1_mask) at (14,6,\Z) [cls, fill=\C!30] {};
\node (patch_2_mask) at (14,5.5,\Z) [cls, fill=\C!30] {};
\node (dots) at (14,4.9,\Z) [] {$\vdots$};
\node (patch_5_mask) at (14,4,\Z) [cls, fill=\C!30] {};
\node (patch_6_mask) at (14,3.5,\Z) [cls, fill=\C!30] {};
\node (patch_7_mask) at (14,3,\Z) [cls, fill=\C!30] {};
}

\node (out_patch_first) at (16,6) [] {\includegraphics[height=41.25pt]{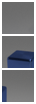}};
\node (dots) at (16,4.9) [] {$\vdots$};
\node (out_patch_first) at (16,3.5) [] {\includegraphics[height=41.25pt]{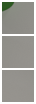}};

\node (output) [right of=dots, anchor=west] {\includegraphics[width=60pt]{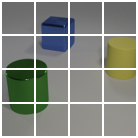}};

\draw [->, thick](0.1,4.75) -- (0.6,4.75);
\draw [-to, thick](1.5,4.75) -- (1.75,4.75);
\draw [-to, thick](1.5,8.25) -- (1.75,8.25);

\draw [-to, thick](3.25,4.75) -- (3.5,4.75);
\draw [-to, thick](3.25,8.25) -- (3.5,8.25);

\draw [-to, thick](4.5,8.25) -| (extraction.east);
\draw [-to, thick](4.5,4.75) -- (4.75,4.75);

\draw [-to, thick](6.25,4.75) -- (6.5,4.75);

\draw [-to, thick](7.5,4.75) -- (7.75,4.75);
\draw [-to, thick](extraction.west) -- ++(0,-0.5) -| (forming.west);

\draw [-to, thick](9.25,4.75) -- (9.5,4.75);
\draw [-to, thick](11,4.75) -- (11.25,4.75);
\draw [-to, thick,red](13.25,4.75) -- (13.5,4.75);

\draw [-to, thick](15,4.75) -- (15.5,4.75);
\draw [-to, thick](16.5,4.75) -- (17,4.75);

\end{tikzpicture}

\caption[The proposed architecture]{The architecture of the transformer-based
autoencoder with the object-centric representations in the bottleneck.}
  \label{fig:architecture}
\end{figure}
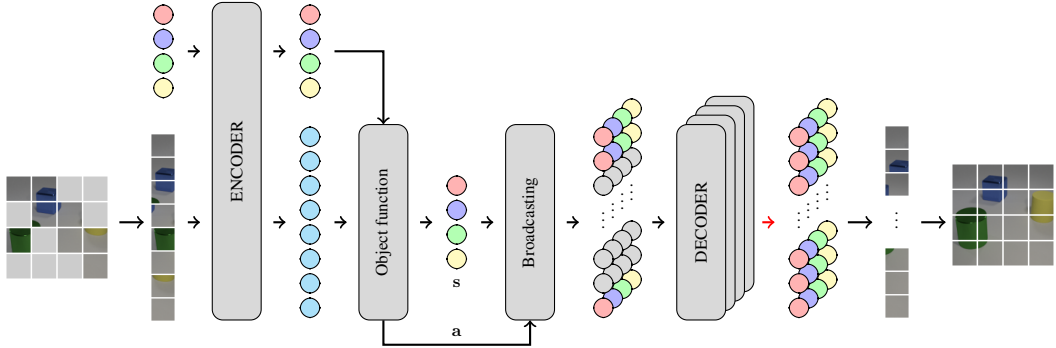

Our model is inspired by the transformer-based masked autoencoder \cite{mae} with modifications to enable end-to-end unsupervised multi-object segmentation and representation learning. The autoencoder receives an image with some of the patches masked out and it is tasked to reconstruct all the patches including both the unmasked and masked ones (see Fig.~\ref{fig:architecture}).

\begin{figure}[t]
  \centering
\subfloat[Object function]{
\begin{tikzpicture}[scale=0.65, every node/.style={scale=0.65}]
  
    \node (cls_0_enc) at (0,6) [cls, fill=red!30] {};
    \node (cls_1_enc) at (0,5.5) [cls, fill=blue!30] {};
    \node (cls_2_enc) at (0,5) [cls, fill=green!30] {};
    \node (cls_3_enc) at (0,4.5) [cls, fill=yellow!30] {};
  
    \node (patch_0_enc) at (0,3.5) [patch] {};
    \node (patch_1_enc) at (0,3) [patch] {};
    \node (patch_2_enc) at (0,2.5) [patch] {};
    \node (patch_3_enc) at (0,2) [patch] {};
    \node (patch_4_enc) at (0,1.5) [patch] {};
    \node (patch_5_enc) at (0,1) [patch] {};
    \node (patch_6_enc) at (0,0.5) [patch] {};
    \node (patch_7_enc) at (0,0) [patch] {};
    
    \node (dot_product) at (2,1.75) [rectangle,rounded corners,text centered, draw=black,fill=gray!30,rotate=90,minimum width=3cm, minimum height=1cm] {Scaled dot product};

    \node (softmax) at (4,1.75) [rectangle,rounded corners,text centered, draw=black,rotate=90,minimum width=3cm, minimum height=1cm,fill=purple!30] {Softmax};
    
    \node (w_mean) at (6,1.75) [rectangle,rounded corners,text centered, draw=black,rotate=90,minimum width=3cm, minimum height=1cm,fill=gray!30] {Weighted mean};

    \draw [-to, thick](0.5,5.25) -| (dot_product.east);
    
    \draw [-to, thick](0.5,1.75) -- (dot_product.north);
    
    \draw [-to, thick](dot_product.south) -- (softmax.north);
    \draw [-to, thick](softmax.south) -- (w_mean.north);
    \draw [-to, thick](5,1.75) |- ++(0,2) -- node [above]{$\va$} ++(2.5,0);
    
    \draw [-to, thick](w_mean.south) -- node [above]{$\vs$} ++(1, 0);
    \draw [-to, thick](1,1.75) |- ++(0,-2) -| (w_mean.west);
\end{tikzpicture}
  \label{fig:object_function}
}
\subfloat[Broadcasting module]{
\begin{tikzpicture}[scale=0.65, every node/.style={scale=0.65}, node distance=20mm]
  
    \node (loga) at (0,2) [] {$\log\mathbf{a}$};
    \node (s) at (0,0) [] {$\vs$};
    
    \node (broadcast) at (1.5,0) [function] {Broadcast};
    \node (concatenate) [function,below of=broadcast] {Concatenate};
    \node (project) [function,below of=concatenate] {Embed};
    \node (add) [function, below of=project] {Add mask tokens};
    \node (unshuffle) [function, below of=add] {Unshuffle};
    
    \foreach \Z/\C in {-1.5/yellow,-1/green,-0.5/blue, 0/red}
    {
    \node (patch_0_mask) at (11,1.5,\Z) [cls, fill=\C!30] {};
    \node (patch_1_mask) at (11,1.,\Z) [cls, fill=\C!30] {};
    \node (patch_2_mask) at (11,.5,\Z) [cls, fill=gray!30] {};
    \node (dots) at (11,0.,\Z) [] {$\vdots$};
    \node (patch_5_mask) at (11,-.5,\Z) [cls, fill=gray!30] {};
    \node (patch_6_mask) at (11,-1,\Z) [cls, fill=gray!30] {};
    \node (patch_7_mask) at (11,-1.5,\Z) [cls, fill=\C!30] {};
    }
    
    \draw [-to, thick](loga) -| (concatenate.east);
    \draw [-to, thick](s) -- (broadcast);
    \draw [-to, thick](broadcast) -- (concatenate);
    \draw [-to, thick](concatenate) -- (project);
    \draw [-to, thick](project) -- (add);
    \draw [-to, thick](add) -- (unshuffle);
    \draw [-to, thick](unshuffle) -- (dots);
\end{tikzpicture}
  \label{fig:broadcasting}
}
     \caption[The proposed object function]{The operations performed in the bottleneck of the autoencoder.}
\end{figure}
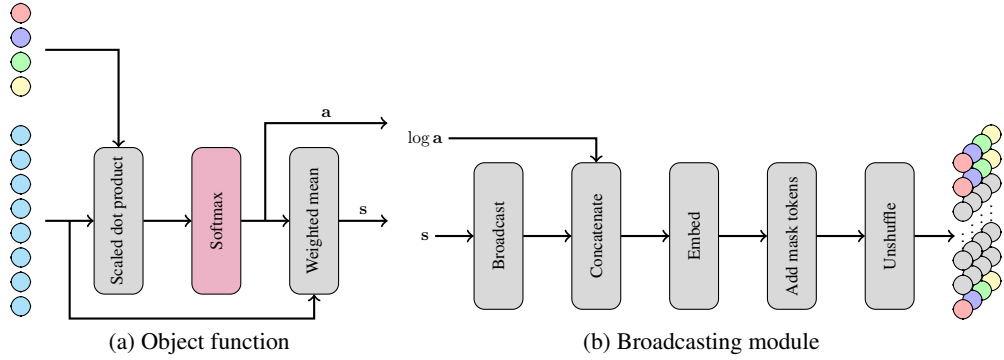

The encoder is a ViT \cite{vit} that receives the unmasked patches together with $K$ learnable embeddings (or class tokens). Just as in MAE \cite{mae}, we add fixed sine-cosine postional encodings from \cite{transformer} to the embeddings of the patches. The output of the transformer encoder is processed by an object function which finds representations of the $K$ individual objects. Inspired by \cite{segmenter}, our object function uses dot product between the encoded class tokens $\mC$ and the encoded unmasked patches $\mZ$ for producing segmentation masks $\mA$ over the patches:
\begin{equation}
\label{eq:patch_mask}
    \mA = \text{Softmax}\left(\frac{\mZ \cdot \mC^{\mathsf{T}}}{\sqrt{D_\text{encoder}}}\right) \in \mathbb{R}^{N_\text{unmasked} \times K}
\,.
\end{equation}
Object representations (or slots) are computed as weighted averages of the encoded unmasked patches $\vz_i$:
\begin{equation}
\label{eq:object_pooling}
    \vs_k = \sum_{i=1}^{N_\text{unmasked}} w_{ik} \vz_i
, \qquad
w_{ik} = \frac{a_{ik}}{\sum_{i=1}^{N_\text{unmasked}} a_{ik} + \epsilon}
\end{equation}
where the weights $w_{ik}$ are computed from the elements $a_{ik}$ of the segmentation masks $\mA$. The object function is schematically visualized in Fig.~\ref{fig:object_function}. 

The decoding procedure gets the slots $\vs$ and the patch masks $\va$ as input and produces $K$ individual reconstructions of the original image and the corresponding $K$ output masks (alpha channels). We use a ViT decoder with two layers to reconstruct the whole image and therefore we form $N=HW/P^2$ inputs using a specially designed broadcasting module (see Fig.~\ref{fig:broadcasting}). Our broadcasting module takes inspiration both from the MAE decoder \cite{mae} and from the spatial broadcast decoder \cite{broadcast_decoder} used in several object-learning models \cite{monet, iodine,slot_attention,genesis, genesisv2, emorl}. In particular, each object representation is repeated $N_\text{unmasked}$ times, where $N_\text{unmasked}$ is the number of the unmasked patches, and these sequences are then concatenated with the respective log-attention masks $\log(\va)$ produced by the object function. 
Then, exactly as in MAE \cite{mae}, we add a learnable mask token at the positions of the masked tokens and the sine-cosine position embeddings. These operations are represented with the 'Unshuffle' block in Fig.~\ref{fig:broadcasting}. 

One problem that we need to address is the difficulty of the masked autoencoding task when we represent the input as a very limited set of object-centric representations. Our model has a narrower bottleneck in comparison to MAE \cite{mae}. Another difficulty is how to form the inputs of the decoder in a differentiable way. To address these problems, we do not only pass the object representations to the decoder, but also the positional information about the unmasked patches used to construct those representations. This positional information is passed in the form of the log-attention masks $\log(\va)$.

The final reconstruction is computed as the sum of the individual reconstructions produced by the slots weighted by their corresponding masks:
\begin{equation}
\label{eq:reconstruction}
    \mathbf{\hat{x}} = \sum_{k=1}^{K} \mathbf{m}_k \mathbf{\hat{x}}_k \in \mathbb{R}^{H \times W \times C}
,
\end{equation}
where $\mathbf{\hat{x}}_k$ is the reconstruction produced by slot $k$ and the output masks $\mathbf{m}_k$ are computed as softmax-normalized alpha channels produced by the ViT decoder.

\subsection{Learning Objective}

The proposed model is an autoencoder and its main objective is to minimize the reconstruction loss:
\begin{equation}
\label{eq:recon_loss}
    \mathcal{L}_\text{rec} = \frac{1}{HWC}\sum_{i=1}^H \sum_{j=1}^W \sum_{c=1}^C (\hat{x}_{ijc} - x_{ijc})^2
.
\end{equation}
In order to facilitate the goal of getting a segmentation of the scenes and having explicit object-centric representations, we similarly to \cite{ast} and \cite{motion_grouping}, introduce an entropy loss on the output masks over the pixels:
\begin{equation}
\label{eq:pixel_entropy}
    \mathcal{L}_\text{pixel} = -\frac{1}{HW}\sum_{i=1}^H \sum_{j=1}^W \left(\sum_{k=1}^{K} m_{kij} \log m_{kij}\right)
.
\end{equation}
This pixel entropy loss encourages the output masks to be near one-hot encoded. The authors of \cite{ast} use the square of the entropies as the loss to avoid problems with singular gradients when masks are near one-hot. We found that our model appeared easier to train using simply the entropy just as done in \cite{motion_grouping}.

While the reconstruction loss combined with the pixel entropy loss was enough to produce good segmentation results on simple datasets, on more difficult data, particularly with very varying backgrounds and numbers of objects, the model could start to split objects or use otherwise object-focused slots for reconstructing part of the background. To this end, we adapt a version of the object entropy loss suggested in \cite{ast}:
\begin{equation}
\label{eq:object_entropy}
    \mathcal{L}_\text{object} = -\sum_{i=1}^{K} \overline{\mathbf{m}}_k\log \overline{\mathbf{m}}_k
\,,
\qquad
    \overline{\mathbf{m}}_k =  \frac{1}{HW}\sum_{i=1}^H\sum_{j=1}^W m_{kij} 
\,.
\end{equation}
The main motivation behind the loss is that if no object is present, then everything in the image is background and the entropy of the averages of the output masks should be one-hot, with the background mask being one \cite{ast}. 
Therefore, introducing such a loss can function as a sort of sparsity inducing penalty. 

The full loss is a weighted sum of the three losses:
\begin{equation}
\label{eq:full_loss}
    \mathcal{L}= \mathcal{L}_\text{rec} + \lambda_\text{pixel} \mathcal{L}_\text{pixel} + \lambda_\text{object} \mathcal{L}_\text{object}.
\end{equation}

\subsection{Training strategy}  

\paragraph{Class token initialization} We learned that larger initial values of the class tokens, most likely in combination with the masking approach, sometimes could lead to class tokens specializing to trivial solutions very early in training. Therefore,
we initialize the class token weights using very small values which are drawn from the zero mean Gaussian distribution with standard deviation of $\sigma=0.002$. 

\paragraph{Masking strategy} 
There are potential challenges with using masking for object-centric learning. Firstly, the choice of masking strategy and masking ratio could impact the learning. As we in this work consider multi-object scenes, a too high masking ratio or, for example, block-wise masking strategies can fully mask out some or several objects. This could make the training unstable. We therefore choose the random masking strategy, exactly as in MAE \cite{mae}. 

We use the masked autoencoding task to encourage the model to learn meaningful object-centric representations. However, our end goal is to extract object representations from full images. Therefore, we adjust the masking strategy during training such that we start with a relatively large ratio of masked patches and keep it fixed during the initial period of the learning rate warm-up. After that, we linearly decrease this ratio towards zero during the course of training.

\paragraph{Weighting of the loss terms} We note that there is some degree of sensitivity to the weights of the loss terms in \eqref{eq:full_loss}. We therefore adopt a strategy similar to the one of the annealing the masking ratio. The loss weights $\lambda_\text{pixel}$, $\lambda_\text{object}$ are kept fixed to relatively small values during the learning rate warmup period. After this, these weights are increased linearly to reach their final value in the last epoch. 

\section{Experiments}
\label{sec:experiments}

\textbf{Datasets} \ The model is trained and evaluated on four multi-object datasets commonly used as benchmarks for object-centric representation learning. The first three of these are Tetrominoes, gray scale background multi-dSprites and CLEVR6 provided in \cite{multiobjectdatasets19}. In particular, we use the processed versions of these that were used in \cite{emorl}\footnote{Accessed from: \url{https://doi.org/10.5281/zenodo.4895643}} for efficient usage with PyTorch. For CLEVR6, we crop and resize images to $128 \times 128$ resolution similarly to \cite{iodine, slot_attention, dti}. 

We follow the procedures of previous work and use 60K training images and 320 evaluation images for Tetrominoes and multi-dSprites \cite{iodine, slot_attention, emorl, dti}. For CLEVR6, as in \cite{emorl}, we have 50K training images and also hold out 320 images for evaluation. We run experiments for five random seeds for each of these datasets similarly to previous works \cite{iodine, slot_attention, emorl, dti}. 

We note that the authors of \cite{dti} report that the CLEVR6 dataset includes roughly 35K images, while we use the version of CLEVR6 from \cite{emorl} which has 50K images. Thus, there might be some differences in the amount of training data for CLEVR6 between the reported results. However, given the simple nature of the dataset, we can still make a relatively fair comparison.

We also test the proposed model on the ClevrTex dataset and the generalization test datasets OOD and CAMO described in \cite{clevrtex}. For ClevrTex, we use the image preprocessing techniques and the data splits as described in \cite{clevrtex} and implemented in their dataloader code\footnote{Datasets, data loading and evaluation code were accessed from the page linked in \cite{clevrtex}: \url{https://www.robots.ox.ac.uk/~vgg/data/clevrtex/}}. Due to model instability issues, however, for the ClevrTex data, we run experiments for 10 random seeds in contrast to the three used in the experiments in \cite{clevrtex}. 

\textbf{Setup} \ For all the datasets we use a four-layer ViT encoder and a two-layer ViT decoder. The number of the class tokens is dataset-specific so that $K$ is set to the maximum number of objects per scene plus one for background. The patch size is chosen depending on the dataset resolution such that in no dataset we have more than $N=64$ patches in total. For example, on CLEVR6 and ClevrTex, we use the patch size of $16 \times 16$. We present the detailed hyperparameters in Appendix~\ref{sec:experimental_details}. These hyperparameters have been chosen quite arbitrarily based on what initially worked the best without much tuning.

\textbf{Training} \ Training on all the datasets is done from scratch. Since the masking ratio is decreased to zero during training, there is no need for additional finetuning. We train our models with the AdamW optimizer \cite{adamw} using a batch size of 128. We train for 300 epochs plus additional 30 cool-down epochs on all datasets. We use the 30 cool-down epochs where everything is kept constant to mitigate small instabilties of the masking ratio becoming zero in the end. We perform a linear warmup of the learning rate for 10 epochs. After the initial warmup, the learning rate is decayed with a half-cycle cosine \cite{cosine} until the 300th epoch. When training on ClevrTex, we found that there was a strong tendency for the model to start segmenting the image into trivial non-object centric parts early on in training. To alleviate this, we add zero mean gaussian noise with standard deviation $\sigma=0.1$ to the class tokens before concatenating them during the warmup. However, after the warmup we instantly stop this addition, as otherwise the class tokens tended to start overfitting to positions.

\textbf{Metrics} \ For evaluation of the model, we use mainly segmentation metrics calculated for the decoder output masks. Since we use quite large patches, detailed and accurate segmentation masks should be a good indication of good object learning. In particular, we use the adjusted rand index for foreground (ARI-FG), which is a commonly used metric for evaluation of object-centric models \cite{iodine, slot_attention, emorl, dti}, and the adjusted rand index (ARI). For evaluation on ClevrTex , we follow a similar procedure as in \cite{clevrtex} and thus report also the mean intersection over union (mIoU). For metric calculations, we use the evaluation code provided with \cite{clevrtex}.

\subsection{Qualitative Results}
\label{sec:qualitative}

In this section, we investigate the individual reconstructions $\hat{\vx}_k$ and the output segmentation masks $\vm_k$. The images shown were picked at random from the evaluation set. We present additional qualitative results showing some of the failure cases in Appendix~\ref{sec:additional_results}.

In Fig.~\ref{fig:tetrominoes_outputs}, we show the segmentation results for Tetrominoes. The model demonstrates the ability to create very precise output masks for these data. 

\begin{figure}[p]
  \centering
  \includegraphics[height=40mm,trim={0 0 53mm 0 },clip]{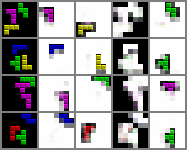}
  \hspace{5mm}
  \includegraphics[height=40mm]{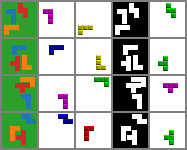}
  \caption[Qualitative evaluation, Tetrominoes]{Qualitative evaluation for the Tetrominoes data. Left: Original images. Right: Predicted segmentations and individual reconstructions with alpha channel.}
  \label{fig:tetrominoes_outputs}
\end{figure}

\begin{figure}[p]
  \centering
  \includegraphics[height=40mm,trim={0 0 140mm 0},clip]{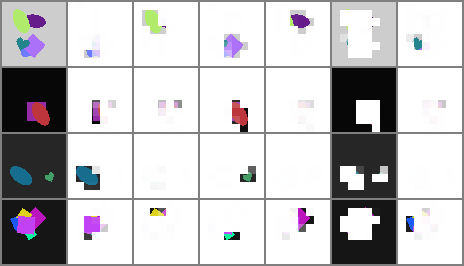}
\hspace{5mm}
  \includegraphics[height=40mm]{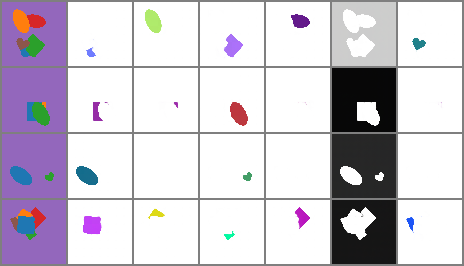}
  \caption[Qualitative evaluation, multi-dSprites]{Qualitative evaluation for the multi-dSprites data. Left: Original images. Right: Predicted segmentations and individual reconstructions with alpha channel.}
  \label{fig:m_dsprites_outputs}
\end{figure}
\begin{figure}[p]
  \centering
  \includegraphics[height=40mm,trim={0 0 321mm 0 },clip]{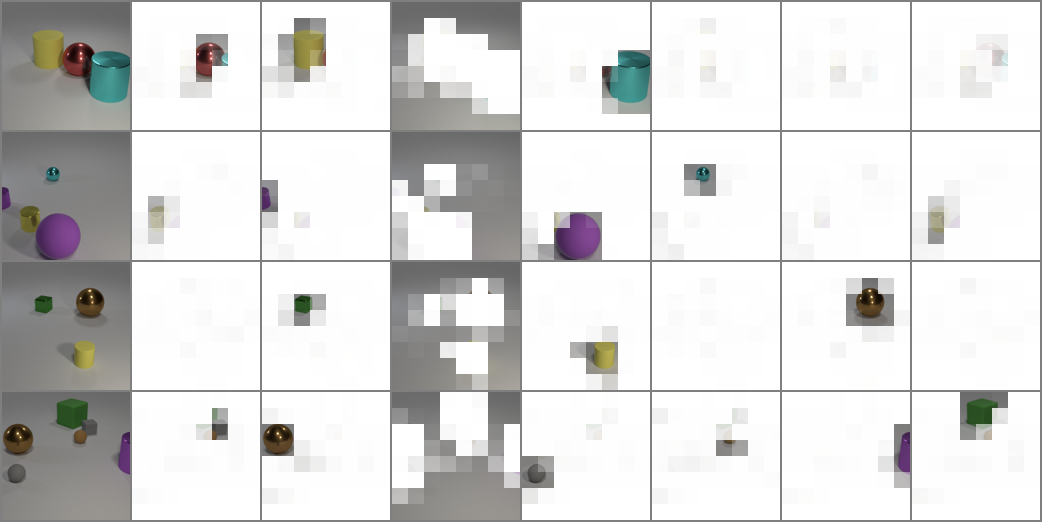}
  \hspace{5mm}
  \includegraphics[height=40mm]{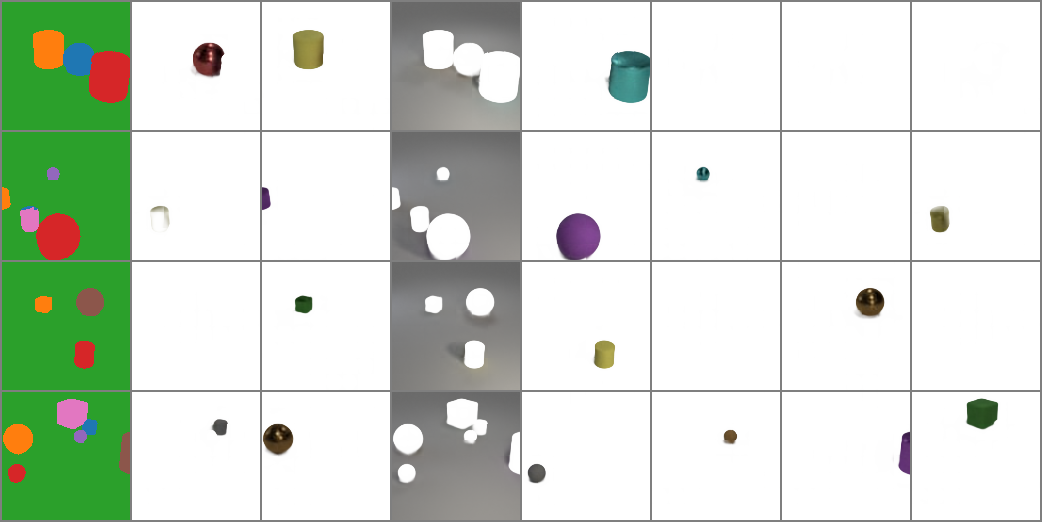}
  \caption{Qualitative evaluation for  the CLEVR6 data. Left: Original images. Right: Predicted segmentations and individual reconstructions with alpha channel.}
  \label{fig:clevr6_outputs}
\end{figure}
\begin{figure}[p]
  \centering
  \includegraphics[height=40mm,trim={0 0 505mm 0},clip]{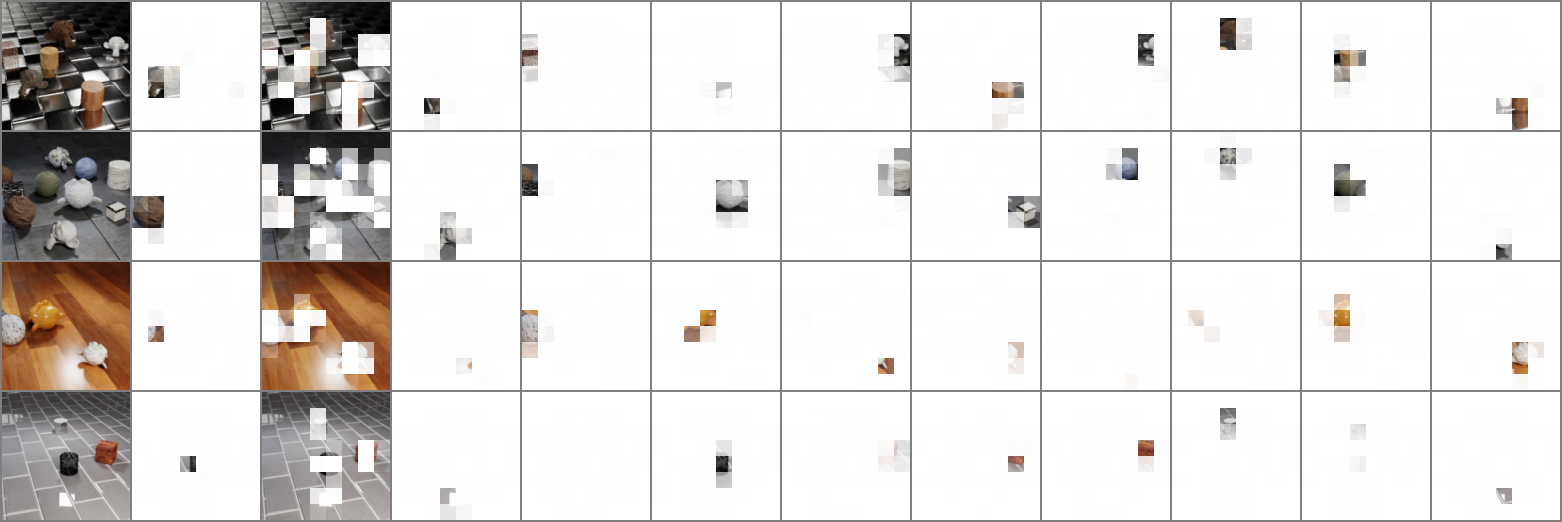}
  \hspace{3mm}
  \includegraphics[height=40mm]{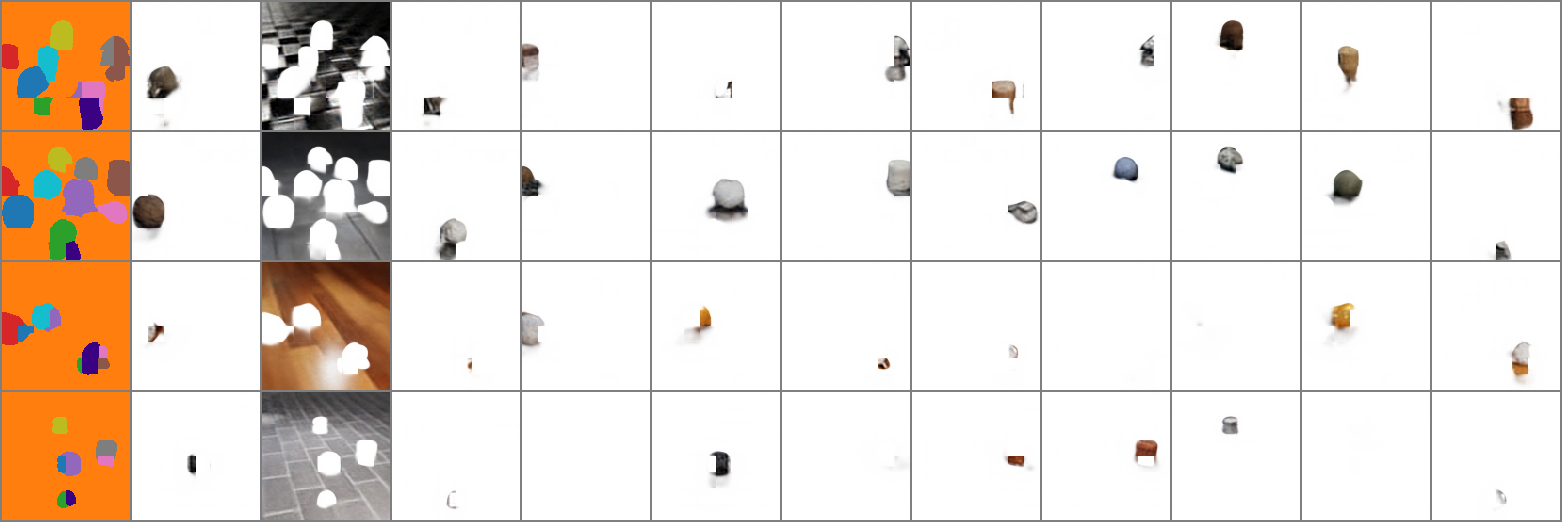}
  \caption[Qualitative evaluation, ClevrTex]{Qualitative evaluation for the ClevrTex test data. Left: Original images. Right: Predicted segmentations and individual reconstructions with alpha channel.}
  \label{fig:clevrtex_outputs}
\end{figure}

In Fig.~\ref{fig:m_dsprites_outputs}, we visualize the same for the multi-dSprites data. The results show that the model can handle cases when objects heavily overlap and when the number of objects is smaller than the number of class tokens used. Especially handling of the overlaps is impressive considering that the model processes patches of size $8 \times 8$.

Fig.~\ref{fig:clevr6_outputs} shows the predicted segmentation for CLEVR6. Despite the very large patches $16 \times 16$, the model does well in discovering small or covered objects and outputs quite precise object masks. However, we note that for CLEVR6 one object in the third row has been reconstructed by two different slots. We also see that the model a bit struggles with the shadows. For the most part, the major shadows are reconstructed with the background but some shadows can be seen in the object reconstructions.

Finally in Fig.~\ref{fig:clevrtex_outputs}, we visualize the segmentations obtained for the ClevrTex test set for one of the three most successful runs. The image samples are chosen randomly. One can clearly see that the class tokens specialize to some spatial positions. Furthermore, we see that the common failure of the model's successful runs is to split objects, so that two slots reconstruct part of the same objects. 

The perhaps most interesting result is how well the model separates the background for all the datasets, even though the background gets no special treatment in our model.

\subsection{Quantitative Results}
\label{sec:quantitaive}

In Table~\ref{tab:arifg_tmc}, we report the ARI-FG for Tetrominoes, multi-dSprites and CLEVR6. Similarly to \cite{iodine, slot_attention, emorl, dti}, the results are averages across five seeds and we also present the standard deviation. For comparison, we present the numbers reported in the respective papers of the other models. For MONet, we use the numbers from \cite{iodine}. Following \cite{dti}, we also report the full ARI which reflects how well the background is segmented.

\begin{table}[h]
	\caption{ARI-FG and ARI means and standard deviations over five seeds for Tetrominoes, Multi-dSprites and CLEVR6. * denotes that one outlier has been filtered.}
	\centering
	\begin{tabular}{llll}
	    \toprule
	     & \textbf{Tetrominoes} & \textbf{M-dSprites} &\textbf{CLEVR6} \\
		\midrule
		\textbf{ARI-FG}\\
		MONet \cite{monet} &  $-$ & $90.4\pm0.8$ & $96.2\pm0.6$ \\
		IODINE \cite{iodine} & $99.2\pm0.4$ & $76.7\pm5.6$ & $98.8\pm0.0$ \\
		SA \cite{slot_attention} & $99.5\pm0.2$* & $91.3\pm0.3$ & $98.8\pm0.3$ \\
		EMORL \cite{emorl} & $98.2\pm1.8$ & $91.2\pm0.4$ & $96.2\pm1.6$ \\
		DTI \cite{dti} & $99.6\pm0.2$ & $92.5\pm0.3$* & $97.2\pm0.2$ \\
		Ours & $99.5\pm0.3$ & $94.7\pm0.4$ & $96.2\pm0.2$ \\
		\midrule
		\textbf{ARI}\\
		DTI \cite{dti} & $99.8\pm0.1$ & $95.1\pm0.1$* & $90.7\pm0.1$ \\
		Ours & $99.9\pm0.0$ & $99.1\pm0.1$ & $97.4\pm0.1$ \\
		\bottomrule
	\end{tabular}
	\label{tab:arifg_tmc}
\end{table}

We note that in particular on multi-dSprites our model compares favourably. On Tetrominoes, our model is more or less on par with the best models and for CLEVR6 the model is slightly worse than the best ones. Most noteworthy is that the segmentation metrics do indicate that the model is able to output accurate segmentations masks despite having a decoder with only two layers and processing only patches of quite large size. Note also that our model is very good at segmenting the background which is indicated by high ARI scores. ARI is not reported for the majority of the models and we believe that our model would compare favourably to many alternative models in this metric. However, we note that we use entropy losses, something many others have not used.

\begin{figure}[t]
	\centering
	\subfloat[mIoU] {
	    \includegraphics[width=150pt]{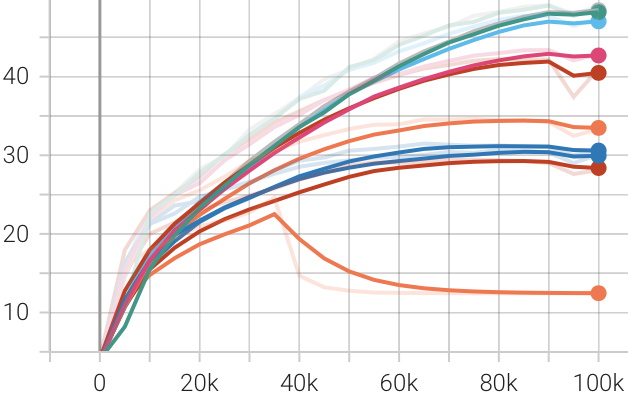}
	}
	\subfloat[ARI-FG] {
	    \includegraphics[width=150pt]{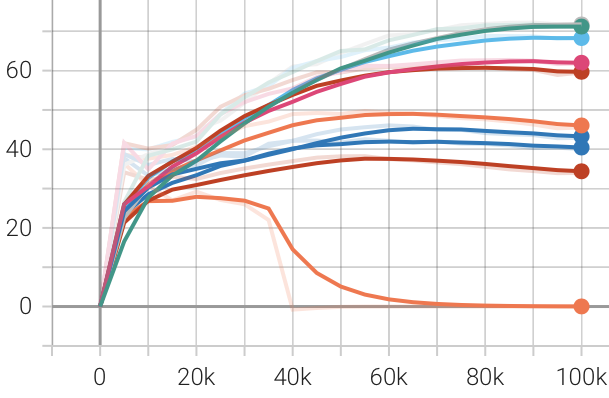}
	}
	\caption[ARI-FG and mIoU development, ClevrTex]{The development of ARI-FG and mIoU on the ClevrTex validation set during training of the base model.}
	\label{fig:clevrtex_validation}
\end{figure}

As the next evaluation of the model, we present metrics for the more difficult ClevrTex in Table~\ref{tab:clevretex_results}. Our model ultimately proved unstable on the ClevrTex data and thus we run it for ten random seeds. We present both average scores computed from the three most successful runs (based on mIoU) and average scores from all ten runs. We investigate the failure cases in Appendix~\ref{sec:additional_results}. To compare our model to others, we report the results from \cite{clevrtex} and add the results from the first version of \cite{ast}\footnote{The second version of \cite{ast} uses a backbone pre-trained with extra data.}. The results for the comparison methods are averages and standard deviations computed from three random seeds. SPAIR* is a modification of the original SPAIR \cite{spair} which contains an additional variational autoencoder for handling background \cite{ast}. For eMORL, we report only the updated results from \cite{clevrtex}. The table shows that the results obtained with our model vary much across different runs. For the three best seeds, the results are fairly consistent and compare well to other methods. In Fig.~\ref{fig:clevrtex_validation}, we show the evolution of the ARI-FG and mIoU scores on the ClevrTex validation set during training of our model. One can see that the results are not stable with one of the seeds failing completely.

\begin{table}[t]
	\caption{ARI-FG and mIoU comparison on ClevrTex test data. The table is from \cite{clevrtex} with added information about AST \cite{ast}.}
	\centering
	\begin{tabular}{lcc}
		\toprule
		& \textbf{mIoU} & \textbf{ARI-FG}  \\
		\midrule
		SPAIR* \cite{spair} & $0.00\pm0.00$ & $0.00\pm0.00$ \\
		SPACE \cite{space} & $9.14\pm3.46$ & $17.53\pm4.13$  \\
		GNM \cite{gnm} & $42.25\pm0.18$ & $53.37\pm0.67$  \\
		MN \cite{mn} & $10.46\pm0.10$ & $38.31\pm0.70$ \\
		DTI \cite{dti} & $33.79\pm1.30$ & $79.90\pm1.37$ \\
		GenV2 \cite{genesisv2} & $7.93\pm1.53$ & $31.19\pm12.41$\\
		eMORL \cite{emorl} & $30.17\pm2.60$ & $-$ \\
		MONet \cite{monet} & $19.78\pm1.02$ & $36.66\pm0.87$ \\
		SA \cite{slot_attention} & $22.58\pm2.07$ & $62.40\pm2.23$ \\
		IODINE \cite{iodine} & $29.16\pm0.75$ & $59.52\pm2.20$  \\
		AST \cite{ast} & $66.62\pm0.80$ & $89.11\pm1.15$ \\
		\midrule
		Ours (top 3) & $48.34\pm0.76$ & $70.23\pm1.80$  \\
		Ours  & $36.40\pm11.65$ & $49.51\pm22.02$ \\
		\bottomrule
	\end{tabular}
	\label{tab:clevretex_results}
\end{table}

We further evaluate the generalization ability of the model with the provided ClevrTex CAMO and OOD datasets as suggested in \cite{clevrtex}. Using our top three models from ClevrTex evaluation, the model achieved an ARI-FG of $67.74\pm1.87$ and an mIoU of $39.80\pm1.14$ on the OOD data. The top three on the CAMO data, get ARI-FG of $54.61\pm1.23$ and mIoU of $42.94\pm0.66$. However, when accounting for all random seeds, ARI-FG and mIoU become $48.99\pm21.03$ and $30.57\pm9.10$ for OOD, and $38.83\pm17.22$ and $33.10\pm9.82$ for CAMO.

Finally, we note that the proposed model is relatively light weight and can be trained pretty fast. Despite using a relatively large batch size of 128 we can train the model on a 16GB GPU. To obtain the reported results, we trained our model for relatively few iterations relatively quickly (see Table~\ref{tab:resource_requirements}). 

\begin{table}[h]
	\caption[Resource requirements for training]{Approximate resource requirements for training with a 16GB V100 GPU. The times include evaluations during and after training.}
	\centering
	\begin{tabular}{llll}
	\toprule
	& \textbf{Iterations} & \textbf{Time} \\
	\midrule
	Tetrominoes & $\sim150$K & $\sim1.75$h  \\
	Multi-dSprites & $\sim150$K &  $\sim4.5$h  \\
	Clevr6 & $\sim130$K & $\sim10.5$h  \\
	ClevrTex & $\sim100$k & $\sim15$h 
	\end{tabular}
	\label{tab:resource_requirements}
\end{table}

\section{Conclusions}

In this paper, we propose a transformed-based autoencoder for explicit learning of object-centric representations. We train the model end-to-end on the task of masked autoencoding and show that the proposed model achieves promising results on a variety of standard benchmarks. Our ablation studies indicate that masking of the input is crucial for making our model work consistently across datasets. In fact, for datasets like CLEVR6, the model fails completely without the usage of masking during training. 

We note that our model has a relatively simple architecture containing (in contrast to many existing models) no iterative procedure during inference. Despite that, the segmentation results obtained on many standard benchmarks are relatively good. Although this result might indicate that the existing synthetic benchmarks are too simple, we still believe that the task of learning consistent representations from inputs corrupted by masking can help develop useful object-centric representations. Combining this learning principle with other ideas from the self-supervised learning literature is a promising direction for future research.

\begin{ack}
We acknowledge the computational resources provided by the
Aalto Science-IT project and CSC (IT Center for Science, Finland).
We acknowledge the Academy of Finland for the support within the Flagship programme: Finnish Center for Artificial Intelligence (FCAI).
\end{ack}

\bibliographystyle{unsrtnat}
\bibliography{references}

\appendix

\section{Appendix}

\subsection{Additional Results}
\label{sec:additional_results}

\begin{figure}[h]
  \centering
  \includegraphics[height=20mm]{model_outputs/tetrominoes_attention.png}
  \hspace{5mm}
  \includegraphics[height=20mm]{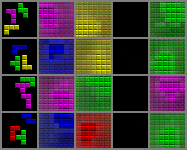}
  \hspace{5mm}
  \includegraphics[height=20mm]{model_outputs/tetrominoes_recons.png} \\ \vspace{5mm}
  \includegraphics[height=20mm]{model_outputs/m_dsprites_attention.png}
  \hspace{5mm}
  \includegraphics[height=20mm]{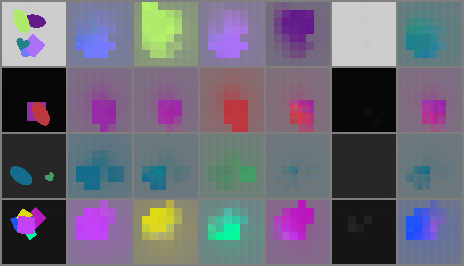}
  \hspace{5mm}
  \includegraphics[height=20mm]{model_outputs/m_dsprites_recons.png} \\ \vspace{5mm}
  \includegraphics[height=20mm]{model_outputs/clevr6_attention.png}
  \hspace{5mm}
  \includegraphics[height=20mm]{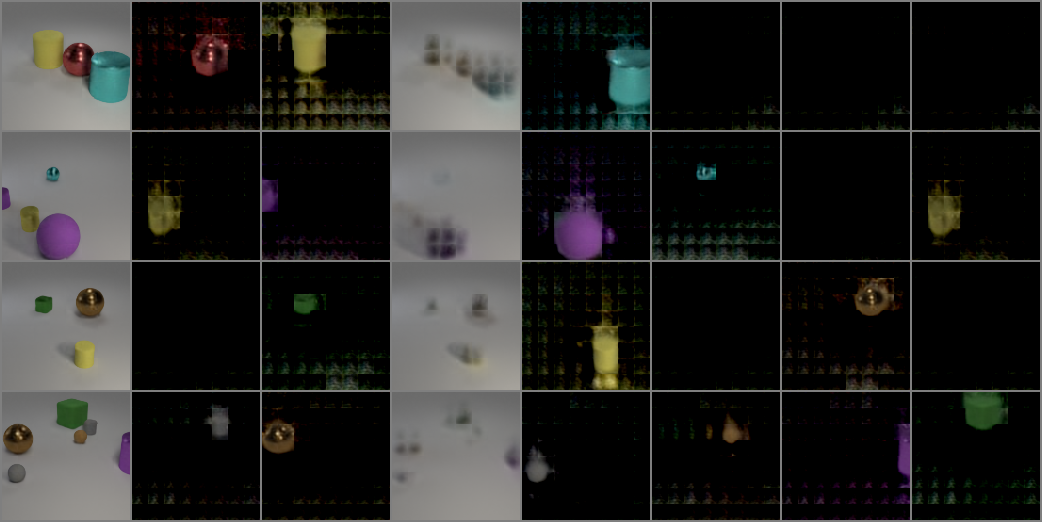}
  \hspace{5mm}
  \includegraphics[height=20mm]{model_outputs/clevr6_recons.png}
  \caption{Qualitative evaluation for the Tetrominoes, multi-dSprites and CLEVR6 data. Left: Original images and bottleneck attention masks. Middle: Full reconstructions and individual reconstructions without alpha channel. Right: Predicted segmentations and individual reconstructions with alpha channel.}
  \label{fig:tmc_appendix}
\end{figure}

\begin{figure}[h]
  \centering
  \includegraphics[height=20mm]{model_outputs/clevrtex_attention.png}
  \hspace{5mm}
  \includegraphics[height=20mm]{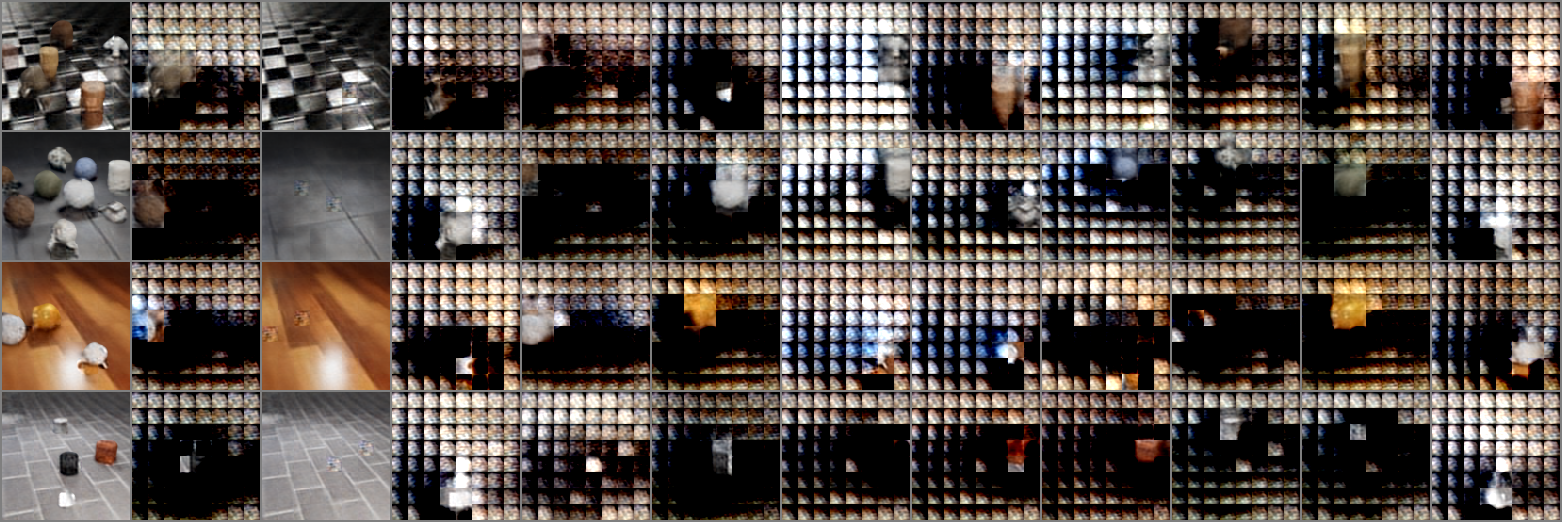}
  \\ \vspace{5mm}
  \includegraphics[height=20mm]{model_outputs/clevrtex_recons.png}
  \caption{Qualitative evaluation for the ClevrTex data. Left: Original images and bottleneck attention masks. Right: Full reconstructions and individual reconstructions without alpha channel. Middle/Below: Predicted segmentations and individual reconstructions with alpha channel.}
  \label{fig:clevrtex_appendix}
\end{figure}

Figures~\ref{fig:tmc_appendix} and \ref{fig:clevrtex_appendix} visualize also the attention masks, full reconstructions and indivdual reconstructions without alpha channel for the examples from \autoref{sec:qualitative}. We note that the shape of the objects are modelled with the alpha channel. Furthermore, we see that while a single patch can contain pieces of multiple objects, the model has found ways to extract information and decode the objects separately for the simpler datasets. For ClevrTex, we clearly see the tendency for class tokens to overfit to certain positions. 

\begin{figure}[h]
  \centering
  \includegraphics[height=40mm,trim={0 0 505mm 0 },clip]{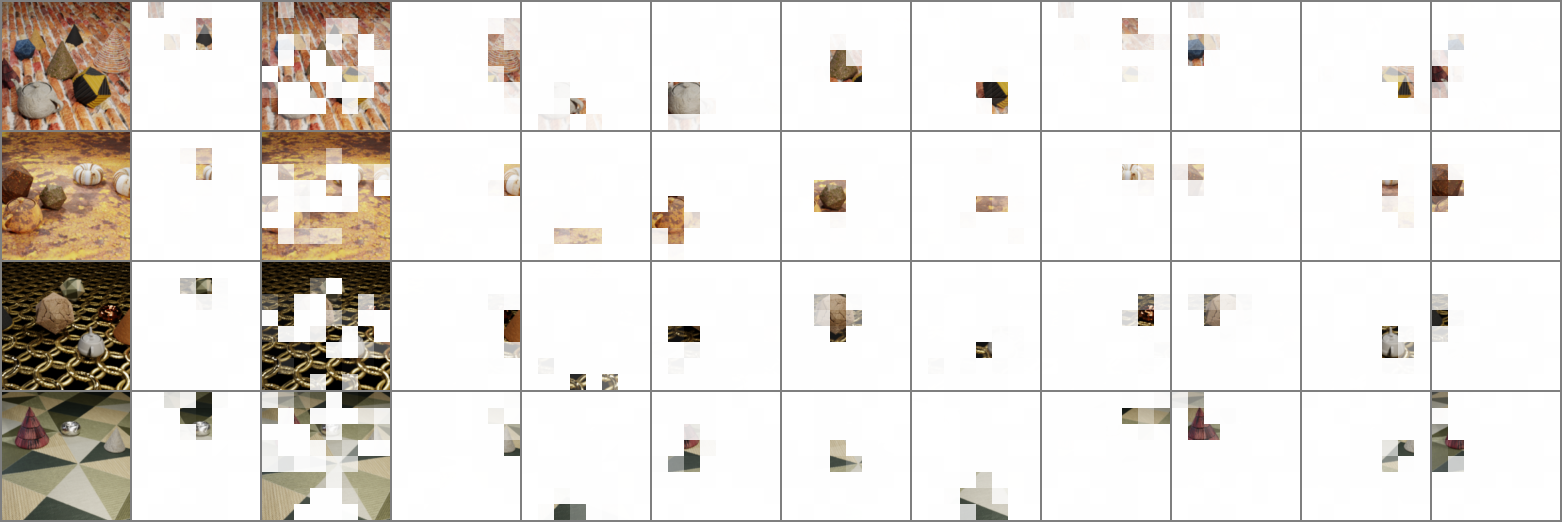}
  \hspace{2mm}
  \includegraphics[height=40mm]{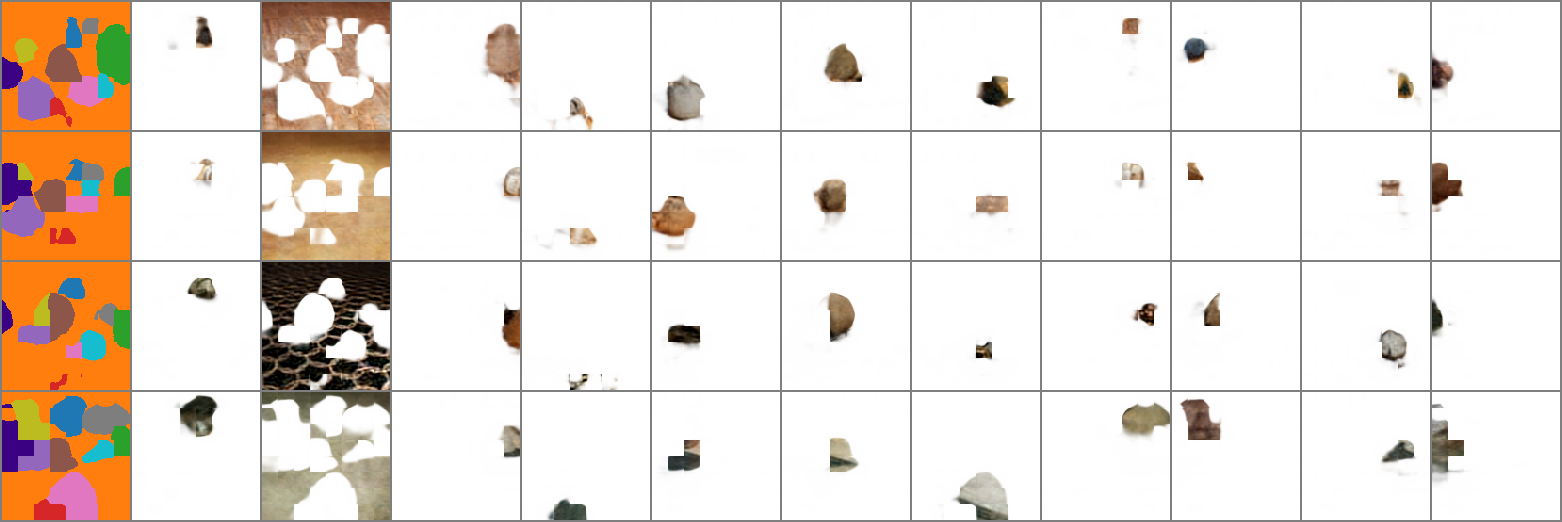}
  \caption{Qualitative evaluation for the ClevrTex OOD data. Left: Original images. Right: Predicted segmentations and individual reconstructions with alpha channel.
  }
  \label{fig:ood_outputs}
\end{figure}

\begin{figure}[h]
  \centering
\includegraphics[height=40mm,trim={0 0 505mm 0 },clip]{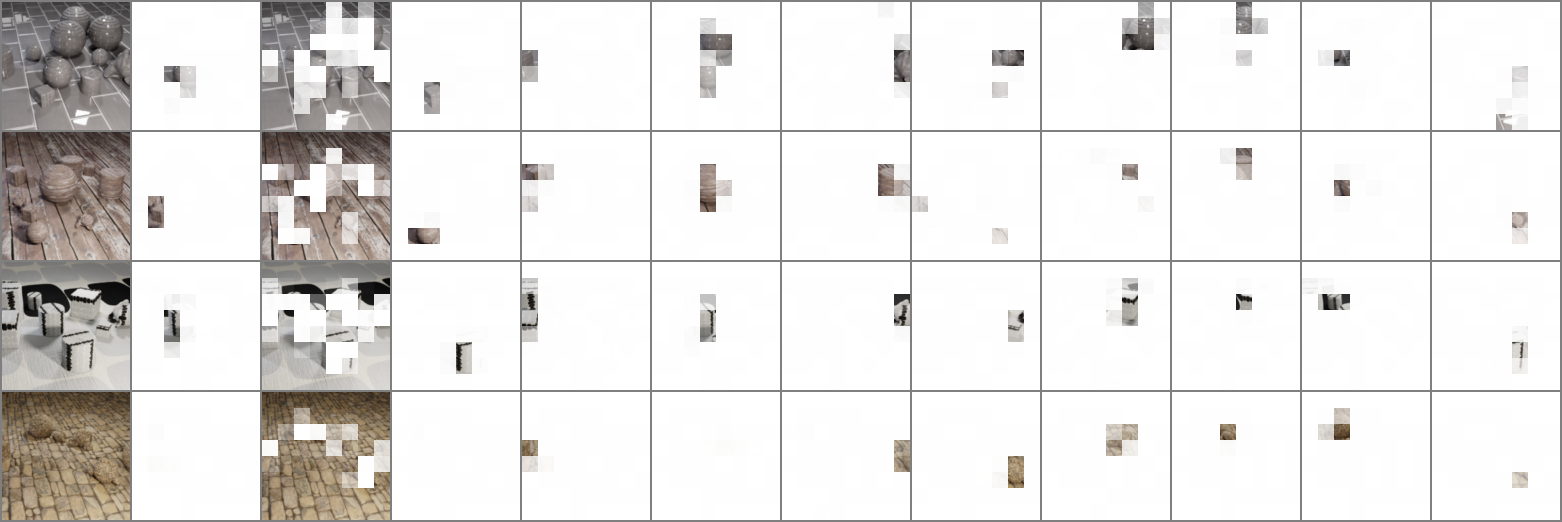}
\hspace{5mm}
\includegraphics[height=40mm]{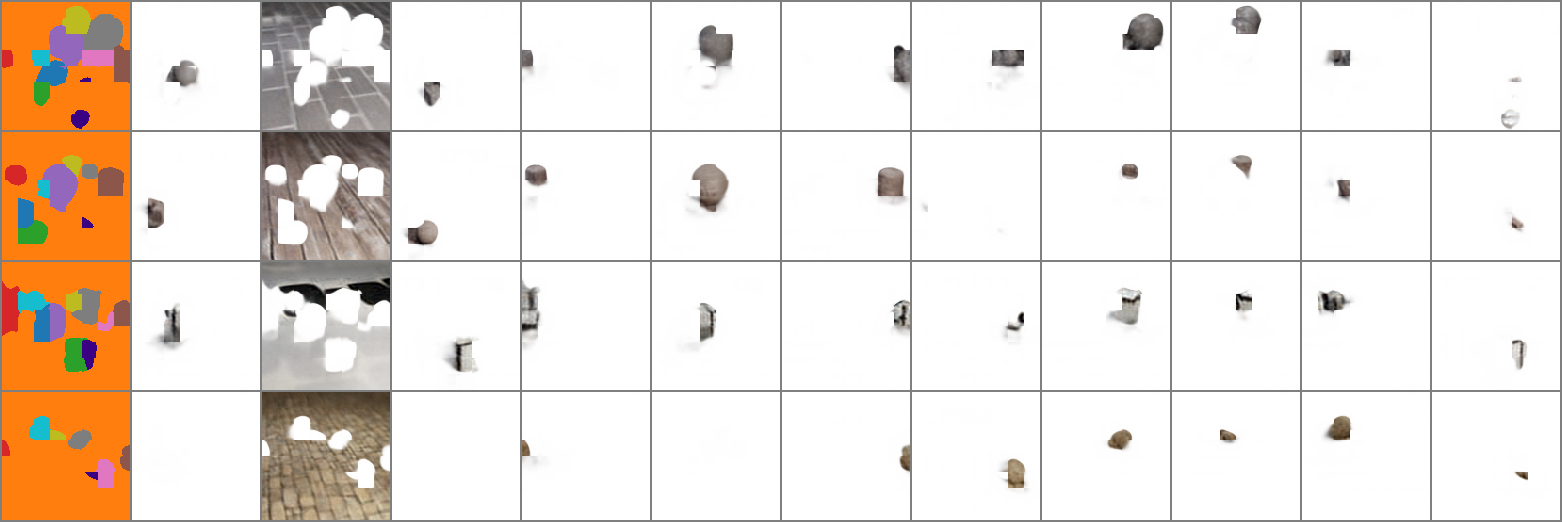}
  \caption{Qualitative evaluation for the ClevrTex CAMO data. Left: Original images. Right: Predicted segmentation and individual reconstructions with alpha channel.
}
  \label{fig:camo_outputs}
\end{figure}

In Figure~\ref{fig:ood_outputs} and \ref{fig:camo_outputs} we present segmentations from also the ClevrTex OOD and CAMO test sets \cite{clevrtex}. We note that the masks are always more or less focused on similar spatial areas, regardless of there being a full object there or not. This often leads to splitting of objects into multiple parts, which was also present on the ClevrTex data, but it is more pronounced for this out of distribution data. We further see that for one of the images in the OOD data, the model has confused parts of the background for objects. Still the background segmentation looks quite good in the most cases on OOD and also for the CAMO data, where objects are harder to separate from background.

\begin{figure}[h]
  \centering
  \includegraphics[width=\textwidth]{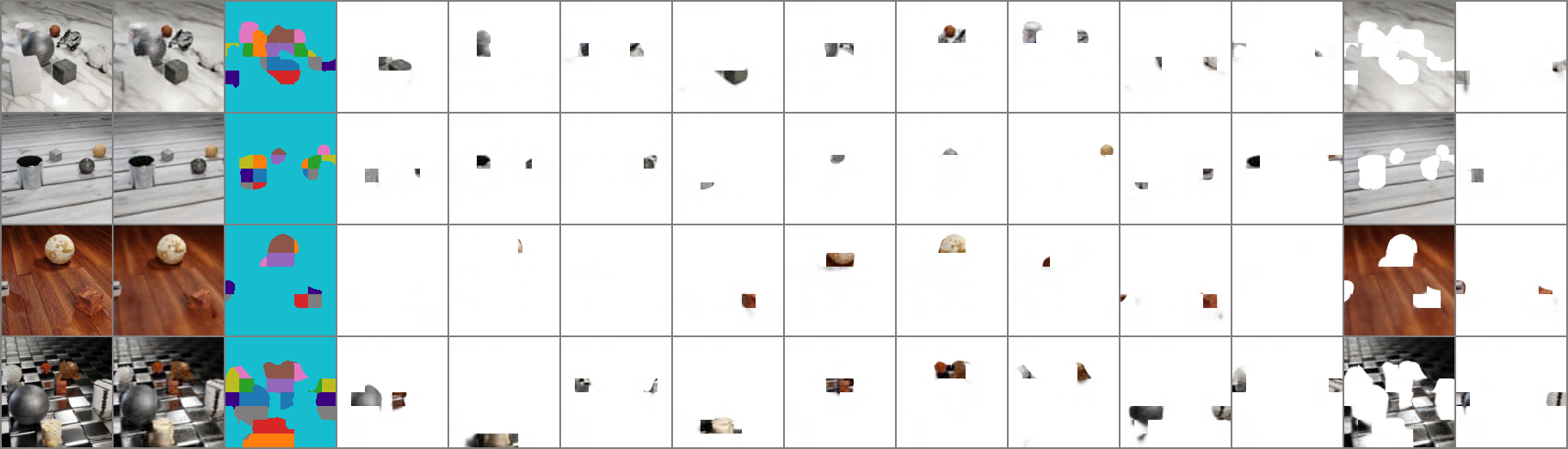}
  \\ \vspace{3mm}
  \includegraphics[width=\textwidth]{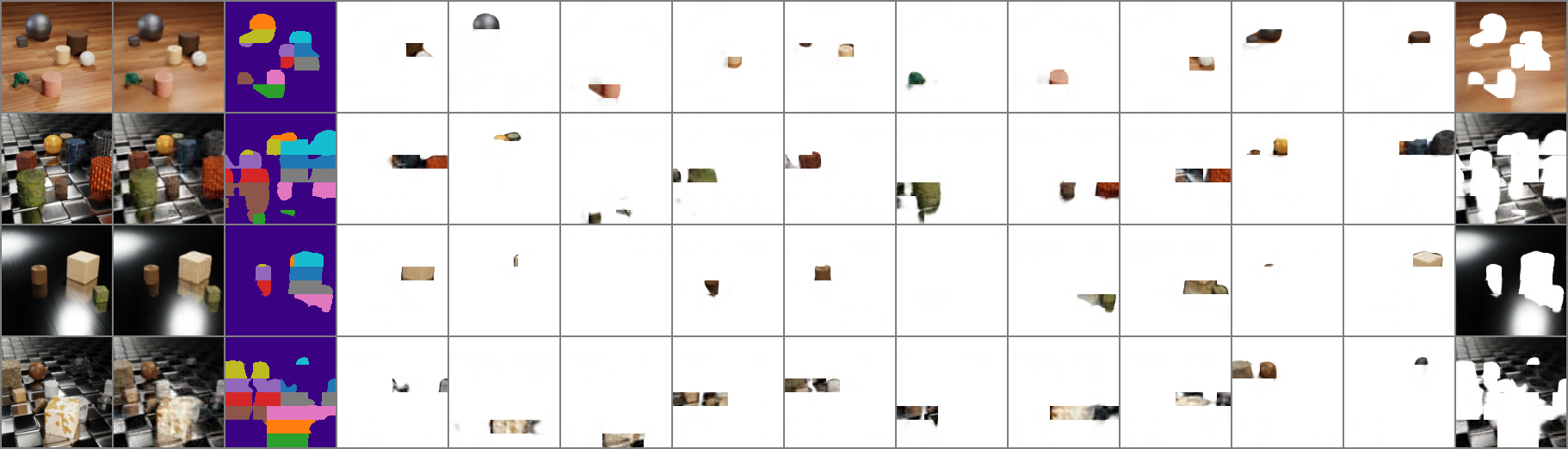}
   \caption{ClevrTex results from failed seeds.}
  \label{fig:clevrtex_failure}
\end{figure}

In \autoref{fig:clevrtex_failure}, we visualize two common failure cases for the model on ClevrTex. We can clearly see that the decomposition of objects does not work, with each slot modelling some horizontal patches regardless of input. However, we see that the background is segmented and reconstructed quite well.

\subsection{Ablations}
\label{sec:ablations}

In this section we present results from varying the training objective. We present ablations for the datasets for which the model was stable, i.e. Tetrominoes, multi-dSprites and CLEVR6. We consider four different cases:

\begin{itemize}
    \item Training without the object entropy loss (w/o obj. ent.).
    \item Training without the pixel entropy loss (w/o pixel. ent.)
    \item Training without any of the two entropy losses (w/o any ent. )
    \item Training without masking (w/o masking)
\end{itemize}

For analysing the differences caused by these changes, we report the  ARI-FG, ARI and mIoU for all of these different cases. The results can be seen in \autoref{tab:ari_fg_ablation}, \ref{tab:ari_ablation} and \ref{tab:miou_ablation}.

\begin{table}[h]
	\caption[Ablations, ARI-FG]{Results on ARI-FG from varying the training objective.}
	\centering
	\begin{tabular}{llll}
	    \toprule
	    & \textbf{Tetrominoes} & \textbf{M-dSprites} & \textbf{CLEVR6} \\
		\midrule
		Base & $99.5\pm0.3$ & $94.7\pm0.4$ & $96.2\pm0.2$ \\
		w/o obj. ent. & $99.5\pm0.5$ & $93.1\pm1.3$ & $98.5\pm0.1$\\
		w/o pixel. ent. & $95.3\pm6.5$ & $75.8\pm41.2$ & $0.0\pm0.0$\\
		w/o any ent. & $99.6\pm0.1$ & $88.1\pm2.7$ & $95.8\pm2.4$\\
		w/o masking & $88.8\pm7.0$ & $94.8\pm0.3$ & $0.0\pm0.0$ \\
		\bottomrule
	\end{tabular}
	\label{tab:ari_fg_ablation}
\end{table}

\begin{table}[h]
	\caption[Ablations, ARI]{Results on ARI from varying the training objective.}
	\centering
	\begin{tabular}{llll}
	    \toprule
	    & \textbf{Tetrominoes} & \textbf{M-dSprites} & \textbf{CLEVR6} \\
		\midrule
		Base & $99.9\pm0.0$ & $99.1\pm0.1$ & $97.4\pm0.1$ \\
		w/o obj. ent. & $99.9\pm0.1$ & $99.0\pm0.2$ & $95.3\pm0.1$\\
		w/o pixel. ent. & $97.4\pm4.0$ & $98.1\pm2.0$ & $0.0\pm0.0$\\
		w/o any ent. & $99.9\pm0.0$ & $73.7\pm34.8$ & $45.1\pm46.4	$\\
		w/o masking & $98.1\pm2.8$ & $98.9\pm0.1$ & $0.0\pm0.0$ \\
		\bottomrule
	\end{tabular}
	\label{tab:ari_ablation}
\end{table}

\begin{table}[!htbp]
	\caption[Ablations, mIoU]{Results on mIoU from varying the training objective.}
	\centering
	\begin{tabular}{llll}
	    \toprule
	    & \textbf{Tetrominoes} & \textbf{M-dSprites} & \textbf{CLEVR6} \\
		\midrule
		Base & $99.7\pm0.2$ & $90.5\pm0.5$ & $75.0\pm0.2$ \\
		w/o obj. ent. & $99.7\pm0.3$ & $87.0\pm2.7$ & $78.4\pm0.3$ \\
		w/o pixel. ent. & $97.2\pm3.9$ & $78.6\pm23.3$ & $17.3\pm0.0$\\
		w/o any ent. & $99.7\pm0.1$ & $72.4\pm8.9$ & $45.6\pm23.2$\\
		w/o masking & $91.3\pm5.9$ & $91.0\pm1.0$ & $17.3\pm0.0$\\
		\bottomrule
	\end{tabular}
	\label{tab:miou_ablation}
\end{table}

It is worth noting that on CLEVR6, the model failed completely without masking. The only dataset where masking appeared irrelevant is multi-dSprites. We hypothesize that this is because the multi-dSprites data mostly includes images were all objects are of different color, and thus it is natural to model the objects separately regardless of whether the training task warrants for it.

Furthermore, the object entropy has a positive effect for multi-dSprites, but for Tetrominoes it has more or less no effect. For CLEVR6, the object entropy has a small negative effect on the segmentation metrics ARI-FG and mIoU. The ARI, however, improves with object entropy for CLEVR6. We still include the object entropy in our base model as it was important for achieving better results on ClevrTex. 

While object entropy for these datasets does not have a significant effect, the pixel entropy does. In particular, for multi-dSprites and CLEVR6, we see that if the pixel entropy is removed, ARI-FG and mIoU are worse on these datasets. In fact, the model with only object entropy fails completely on CLEVR6. If both of the entropy losses are removed simultaneously, the performance on multi-dSprites is worse all around and CLEVR6, especially background segmentation is worse.

\subsection{Experimental Details}
\label{sec:experimental_details}

We present the hyperparameters for the model on the different datasets in \autoref{tab:model_setup}. 

\begin{table}[h]
	\caption[Model setup]{Model hyperparameters for different datasets.}
	\centering
	\begin{tabular}{lllll} 
	    \toprule
	    & \textbf{Tetrominoes} & \textbf{Multi-dSprites} & \textbf{CLEVR6} & \textbf{ClevrTex} \\
	    \midrule
		\multicolumn{5}{l}{\textbf{Embedding}} \\
		Image res. & $35 \times 35$ & $64 \times 64$ & $128 \times 128$ & $128 \times 128$ \\
		Patch size & $5 \times 5$ & $8 \times 8$ & $16 \times 16$ & $16 \times 16$ \\
		Mask ratio & $0.75$ & $0.5$ & $0.75$ & $0.75$ \\
		\midrule
		\multicolumn{5}{l}{\textbf{Encoder}} \\
		Class tokens & $4$ & $6$ & $7$ & $11$\\
		Depth & $4$ & $4$ & $4$ & $4$ \\
		Embed dim. & $192$ & $384$ & $768$ & $768$ \\
		Attention heads & $4$ & $8$ & $16$ & $16$\\
		\midrule
		\multicolumn{5}{l}{\textbf{Decoder}} \\
		Depth & $2$ & $2$ & $2$ & $2$ \\
		Embed dim. & $128$ & $256$ & $512$ & $512$ \\
		Attention heads & $4$ & $8$ & $16$ & $16$ \\
		\bottomrule
	\end{tabular}
	\label{tab:model_setup}
\end{table}

\autoref{tab:loss_weights} presents the initial loss weights during the warmup, and the final weights used during the cooldown. 

\begin{table}[h]
	\caption[Loss function hyperparameters]{Loss function hyperparameters.}
	\centering
	\begin{tabular}{ll}
	    \toprule
	    \textbf{Weight} & \textbf{Value} \\
	    \midrule
	    Init. pixel entropy weight &  $10^{-4}$ \\
		Init. objejct entropy weight & $10^{-4}$ \\
		Final pixel entropy weight &  $3 \cdot 10^{-3}$ \\
		Final object entropy weight & $10^{-2}$ \\
		\bottomrule
	\end{tabular}
	\label{tab:loss_weights}
\end{table}

Then in Table~\ref{tab:training_details} we present the rest of the hyperparameters for training. These are used for all datasets. For learning rate decay we use a half-cycle cosine.

 \begin{table}[h]
	\caption{Training details.}
	\centering
	\begin{tabular}{ll}
	    \toprule
	    \textbf{Hyperparameter} & \textbf{Value} \\
	    \midrule
		Optimizer & AdamW \cite{adamw} \\
		Weight decay & 0.05 \\
		Optimizer momentum & 0.9 \\
		Batch size & 128 \\
		Epochs & 300 \\
		Base learning rate & $5 \cdot 10^{-4}$ \\
		Learning rate warmup & 10 epochs \\
		Start learning rate & $10^{-5}$ \\
		Learning rate schedule & Cosine \cite{cosine} \\
		Minimum learning rate & $10^{-5}$ \\
		Cooldown & 30 epochs \\
		\bottomrule
	\end{tabular}
	\label{tab:training_details}
\end{table}

\end{document}